\documentclass[letterpaper]{article} 
\usepackage[preprint]{aaai2027}  
\usepackage[hyphens]{url}  
\usepackage{graphicx} 
\usepackage{natbib}  
\usepackage{caption} 
\usepackage{algorithm}
\usepackage{algorithmic}
\usepackage{newfloat}
\usepackage{listings}
\DeclareCaptionStyle{ruled}{labelfont=normalfont,labelsep=colon,strut=off} 
\floatstyle{ruled}
\newfloat{listing}{tb}{lst}{}
\floatname{listing}{Listing}
\usepackage{booktabs}
\usepackage{amsmath}   
\usepackage{amssymb}   
\usepackage{multirow}  

\newcommand{\method}{SAPO}                        

\title{SAPO: Self-Supervised Attention Policy Optimization
for Improving Temporal Video Grounding in Vision--Language Models}

\author{
    Zixuan Chen\textsuperscript{\rm 1},\quad
    Depeng Wang\textsuperscript{\rm 2},\quad
    Wei Chen\textsuperscript{\rm 1},\quad
    Ke Xu\textsuperscript{\rm 1},\\
    Ya Guo\textsuperscript{\rm 2},\quad
    Huijia Zhu\textsuperscript{\rm 2}\corresponding,\quad
    Tanfeng Sun\textsuperscript{\rm 1},\quad
    Xinghao Jiang\textsuperscript{\rm 1}\corresponding
}
\affiliations{
    \textsuperscript{\rm 1}Shanghai Jiao Tong University\quad
    \textsuperscript{\rm 2}Ant Group
}

\begin{document}
\maketitle

\begin{abstract}

Video vision--language models (VLMs) have strong semantic understanding, yet
remain unreliable at precise temporal grounding. Existing gains from explicitly
aligning video frames with absolute time suggest that VLMs may already encode
accurate temporal locations internally but lose precision when expressing them
as timestamps. Our analysis across four VLMs and two grounding datasets
confirms this hypothesis: attention intervals outperform predicted intervals
on approximately $66\%$ of valid examples. We term this
discrepancy the \emph{temporal generation gap}. Because precise temporal
annotations are expensive, this finding motivates closing the gap to improve
temporal grounding using only unlabeled video--query pairs. We propose
\textbf{Self-Supervised Attention
Policy Optimization (SAPO)}, which uses internal temporal attention as
self-supervision for Group Relative Policy Optimization and applies a
ground-truth-free criterion to select high-value samples. SAPO consistently
improves temporal grounding across multiple datasets and models spanning weak
to strong initial grounding ability.
Relative mIoU gains reach $33.7\%$ on Qwen2.5-VL-3B and $5.4\%$ on the strong
Time-R1-7B. The improvements transfer across datasets while preserving general
video understanding.

\end{abstract}


\section{Introduction}
Vision--language models (VLMs) have become a general interface for video
understanding \cite{qwen25vl2025}. Beyond recognizing \emph{what} happens, a
capable video VLM must also determine precisely \emph{when} an event occurs.
Temporal video grounding (TVG) directly evaluates this ability by requiring a
model to localize the start and end times of an event described by a
natural-language query. Despite their strong general video understanding,
current VLMs remain substantially less reliable when generating precise
timestamps.

Recent studies attribute much of this weakness to insufficient alignment
between visual features and absolute time. Existing methods therefore inject
frame-level temporal information through interleaved textual timestamps,
timestamp overlays, or time-aligned position embeddings
\cite{timelens2026,vtimellm,numpro,qwen25vl2025}. Their improvements establish
that VLMs predict time more accurately when the correspondence between visual
evidence and temporal coordinates is made explicit.

\begin{figure}[t]
    \centering
    \includegraphics[width=\columnwidth]{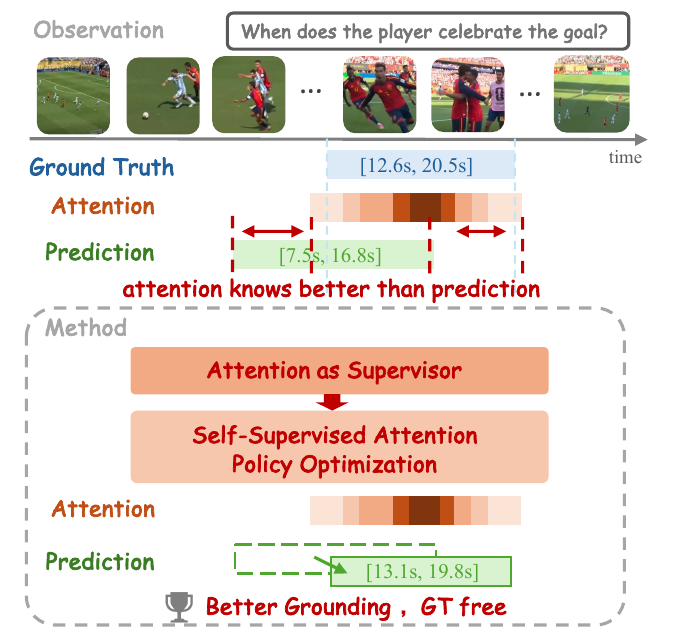}
    \caption{\textbf{Attention knows better than prediction.}
    The attention-decoded interval $I_a$ can localize the queried event more
    accurately than the generated interval $I_p$.
    \method{} uses attention as supervisor to improve native
    timestamp prediction without temporal-boundary annotations during
    training.}
    \label{fig:teaser}
\end{figure}

Given the substantial gains from explicitly mapping visual inputs to temporal
coordinates, we ask whether a VLM already contains internal evidence for
temporal localization but loses precision when generating timestamps. To
investigate this question, we compare the model's explicitly generated
\emph{predicted interval} with an \emph{attention interval} decoded from its
temporal attention over video tokens. We evaluate both intervals by their IoU
with the ground-truth interval and count the proportion of examples for which
the attention interval is more accurate. Across four VLMs, attention achieves a
higher IoU than prediction on an average of $66.0\%$ of examples on
Charades-STA and $65.8\%$ on ActivityNet-Captions. We term this systematic
discrepancy the \emph{temporal generation gap}: the model's attention can
localize the queried event more accurately than its explicit timestamp
prediction, as illustrated in Figure~\ref{fig:teaser}.

Although attention does not provide perfectly accurate boundaries, its relative
advantage reveals a useful source of internal temporal supervision. Since
annotating precise start and end timestamps for video--query pairs is highly
costly, this finding motivates us to exploit unlabeled video--query data and
improve temporal grounding directly through attention-based self-supervision.

Based on this insight, we propose \textbf{Self-Supervised Attention Policy
Optimization (SAPO)}, which uses internal attention to construct RL rewards for
timestamp generation. Each selected video--query pair is processed by a
trainable policy and a frozen reference model initialized from the same VLM.
The policy samples a group of timestamp intervals, while the reference extracts
a stationary temporal-attention distribution. To avoid the sparse supervision
of directly using interval IoU as the reward, SAPO projects each rollout
interval onto the video timeline with a Gaussian kernel and measures its
Wasserstein-1 distance to the reference attention distribution, yielding a
dense temporal reward. Group-relative advantage estimation then updates the policy, with the
reference model also providing KL regularization. SAPO thus transfers internal
temporal evidence into native timestamp prediction without ground-truth
boundaries for training.

Not all video--query pairs, however, contain equally useful attention
supervision. We therefore explore how to identify a higher proportion of
valuable training examples without ground truth. Among the examined criteria,
top-$k$ selection by Jensen--Shannon divergence between the prediction and
attention distributions provides a lightweight solution and retrieves more
high-value examples for RL than random sampling.

We evaluate SAPO on three benchmarks using both Qwen2.5-VL models and
Time-R1-7B, a strong temporal grounding model, without temporal annotations
during training. SAPO yields relative mIoU improvements of up to $33.7\%$ on
Qwen2.5-VL-3B and $5.4\%$ on Time-R1-7B. Gains under cross-dataset training
further demonstrate transfer across data domains, while the Video-MME accuracy
of Qwen2.5-VL-3B remains close to the base model ($58.0$ vs.\ $57.4$),
indicating that SAPO largely preserves general video understanding. Further
ablations verify the effects of high-value sample selection and the individual
training objectives.
Our contributions aresummarized as follows:
\begin{itemize}
    \item Across multiple VLMs and grounding datasets, we identify a
    \emph{temporal generation gap}: internal temporal attention often localizes
    queried events more accurately than explicitly generated timestamps.

    \item We propose \textbf{Self-Supervised Attention Policy Optimization
    (SAPO)}, which uses internal attention to improve temporal grounding through
    reinforcement learning without temporal-boundary annotations.

    \item Experiments across models, datasets, and training sources demonstrate
    consistent grounding improvements for both general-purpose VLMs and a
    strong temporal grounding model, together with cross-dataset transfer and
    largely preserved general video understanding.
\end{itemize}

\section{Related Work}
\noindent\textbf{Vision--Language Models.}
Recent VLMs combine visual encoders with large language models to support
general image and video understanding
\cite{qwen25vl2025,llavaonevision,internvl25}. Although these models exhibit
strong semantic reasoning ability, precise temporal grounding remains
challenging. Our work directly improves the temporal grounding ability of a VLM
without changing its architecture or requiring temporal annotations.

\noindent\textbf{Temporal Video Grounding.}
Temporal video grounding localizes the video segment described by a
natural-language query. Conventional methods learn this task from annotated
temporal intervals \cite{2dtan,momentdetr}. Recent generative VLMs instead
formulate grounding as native timestamp generation. They improve temporal
grounding through architectural designs, temporally grounded instruction data
and supervised fine-tuning
\cite{vtimellm,timechat,hawkeye,timesuite,trace,groundedvideollm,vtgllm,revisionllm},
or more effective post-training algorithms such as reinforcement learning and
on-policy distillation \cite{timer1,videoopd}. Despite their different
approaches to exploiting supervision, these methods rely on annotated temporal
targets.

Training-free methods avoid temporal annotations by constructing external
workflows that match frame-level information from frozen models with the query
\cite{lee2025tag,tfvtg2024,chatvtg}. Although they can produce temporal intervals
without training, they do not improve the native grounding ability of the
underlying VLM. SAPO differs from both supervised and training-free methods: it
uses self-supervision from the model's own attention to improve its native
timestamp generation through policy optimization, without temporal
annotations.

\noindent\textbf{Model Mechanism Analysis.}
Prior studies inspect internal model states to understand how visual evidence is
encoded before the final response
\cite{zhao2024first,Palit_2023_ICCV}. They show that model outputs may not fully
reflect the information available in intermediate representations. Following
this perspective, we analyze temporal attention in generative VLMs and identify
a gap between internal localization and explicit timestamp generation. 

\section{Preliminary}
\label{sec:preliminary}
Given an untrimmed video $V$ and a natural-language query $q$, temporal video
grounding aims to localize the temporal interval corresponding to the queried
event. We distinguish three intervals: the \emph{ground-truth} interval $I_g$
(used only for evaluation and controlled analysis), the \emph{predicted}
interval $I_p$ generated by the model, and the \emph{attention} interval $I_a$
decoded from its temporal attention. For each example, we determine which
interval is more accurate by comparing $\mathrm{IoU}(I_a,I_g)$ with
$\mathrm{IoU}(I_p,I_g)$.

We conduct this analysis with four VLMs on all 3,720 Charades-STA examples and
the same fixed random sample of 4,000 examples from the 17,505 ActivityNet
examples. 
We obtain $I_a$ from the last prompt token's cross-modal attention to the video
tokens. We average over attention heads, pool spatial tokens within each frame to obtain a
normalized temporal profile $a(t)$, where $\sum_t a(t)=1$. We decode this
profile by finding the shortest contiguous frame window whose cumulative
attention mass reaches a threshold $\tau$:
\begin{equation}
    [i^\ast,j^\ast] \;=\; \arg\min_{1\leq i\leq j\leq T} (j-i)
    \quad \text{s.t.}\;\; \sum_{t=i}^{j} a(t) \;\geq\; \tau ,
    \label{eq:decode}
\end{equation}
where $T$ is the number of sampled video frames and $\tau=50\%$. We
then map the frame indices $[i^\ast,j^\ast]$ to their corresponding timestamps
to obtain $I_a$. The model-specific layers, sensitivity to the layer and
threshold, and further extraction details are provided in the Appendix.
Representing both attention and prediction as temporal intervals enables a
direct comparison under the same IoU metric and avoids favoring attention
merely because it is represented as a smooth distribution.
\begin{table}[t]
    \centering
    \small
    \setlength{\tabcolsep}{2pt}
    \begin{tabular}{@{}llccc@{}}
        \toprule
        \textbf{Model} & \textbf{Data} & \textbf{Attn.\ wins$^\dagger$} & Attn.\ mIoU & Pred.\ mIoU \\
        \midrule
        \multirow{2}{*}{Qwen2.5-VL-3B}
            & Charades    & \textbf{65.0} & 39.5 & 36.5 \\
            & ActivityNet & \textbf{71.0} & 32.8 & 24.6 \\[2pt]
        \multirow{2}{*}{Qwen2.5-VL-7B}
            & Charades    & \textbf{62.0} & 46.1 & 43.6 \\
            & ActivityNet & \textbf{67.0} & 32.4 & 27.4 \\[2pt]
        \multirow{2}{*}{InternVL3-2B}
            & Charades    & \textbf{79.0} & 39.2 & 15.2 \\
            & ActivityNet & \textbf{64.0} & 23.9 & 16.9 \\[2pt]
        \multirow{2}{*}{Qwen3-VL-4B}
            & Charades    & \textbf{58.0} & 46.9 & 44.7 \\
            & ActivityNet & \textbf{61.0} & 43.5 & 41.3 \\
        \bottomrule
    \end{tabular}
    \caption{\textbf{The temporal generation gap across models and datasets.}
    Attn.\ wins$^\dagger$ is the fraction of valid samples where
    $\mathrm{IoU}(I_a,I_g) > \mathrm{IoU}(I_p,I_g)$, excluding all-zero ties.All values
    are percentages.}
    \label{tab:temporal_gap}
\end{table}

\begin{figure}[t]
    \centering
    \includegraphics[width=\columnwidth]{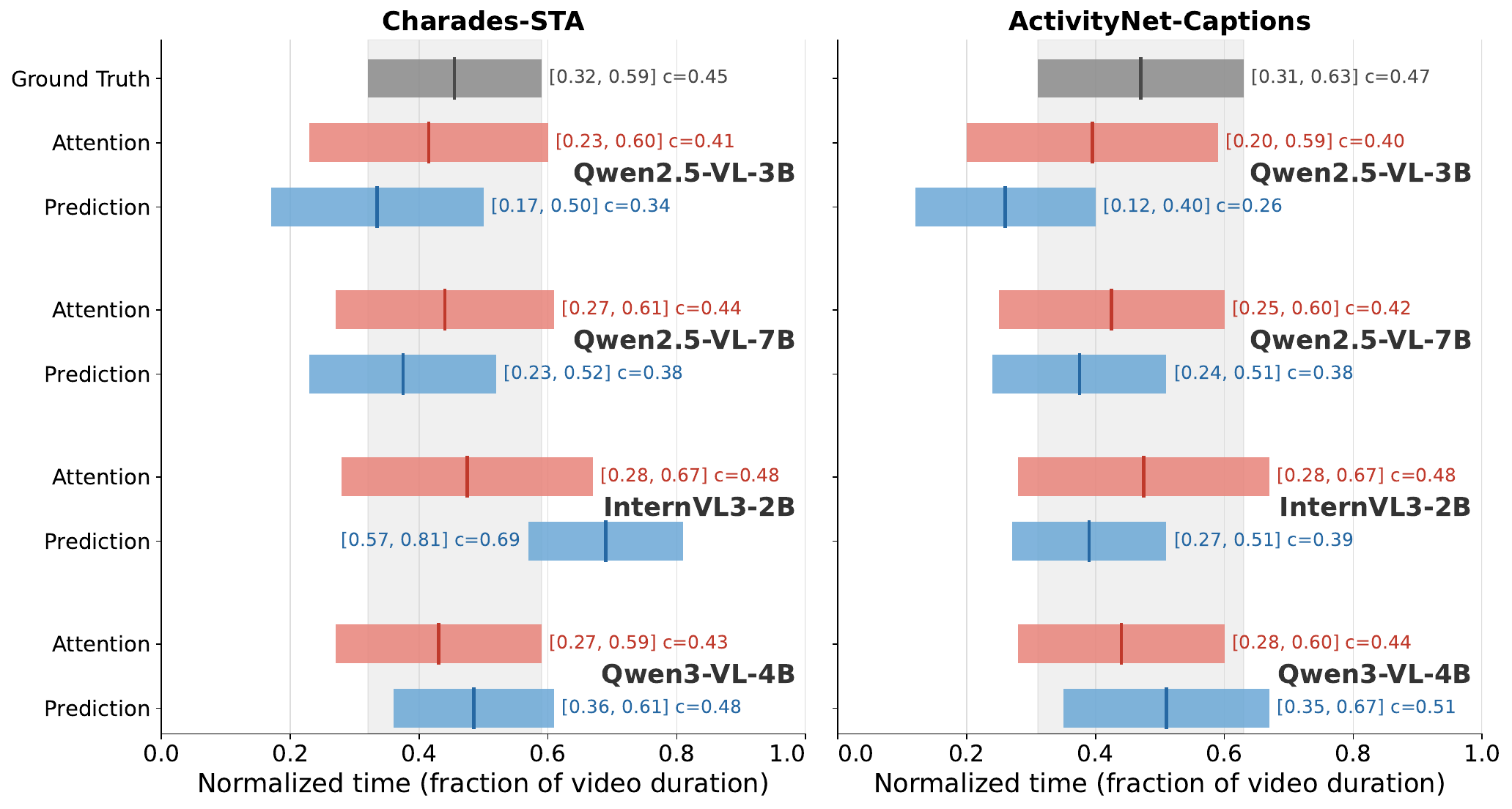}
    \caption{\textbf{Average temporal intervals visualize the generation gap.}
    Each panel overlays the average ground-truth interval, the average predicted
    interval, and the average attention interval for one model--dataset
    pair. The attention interval is consistently closer to the ground-truth
    region than the explicit prediction.}
    \label{fig:avg_intervals}
\end{figure}

Table~\ref{tab:temporal_gap} provides quantitative evidence for a consistent
temporal generation gap. Across all eight model--dataset combinations, the
attention interval achieves higher mIoU than the predicted interval,
and attention is more accurate on the majority of valid examples. The averaged
intervals in Figure~\ref{fig:avg_intervals} exhibit the same pattern. This gap
likely arises because temporal attention operates directly in video-token
position space, whereas timestamp generation must further convert token
positions into calibrated numerical seconds, introducing additional
localization error.
The availability of this effective internal supervision, together with the
high cost of temporal boundary annotation, motivates us to improve timestamp
generation through attention-based self-supervision without human-annotated
boundaries.

\section{Method}
\begin{figure*}[t]
    \centering
    \includegraphics[width=\textwidth]{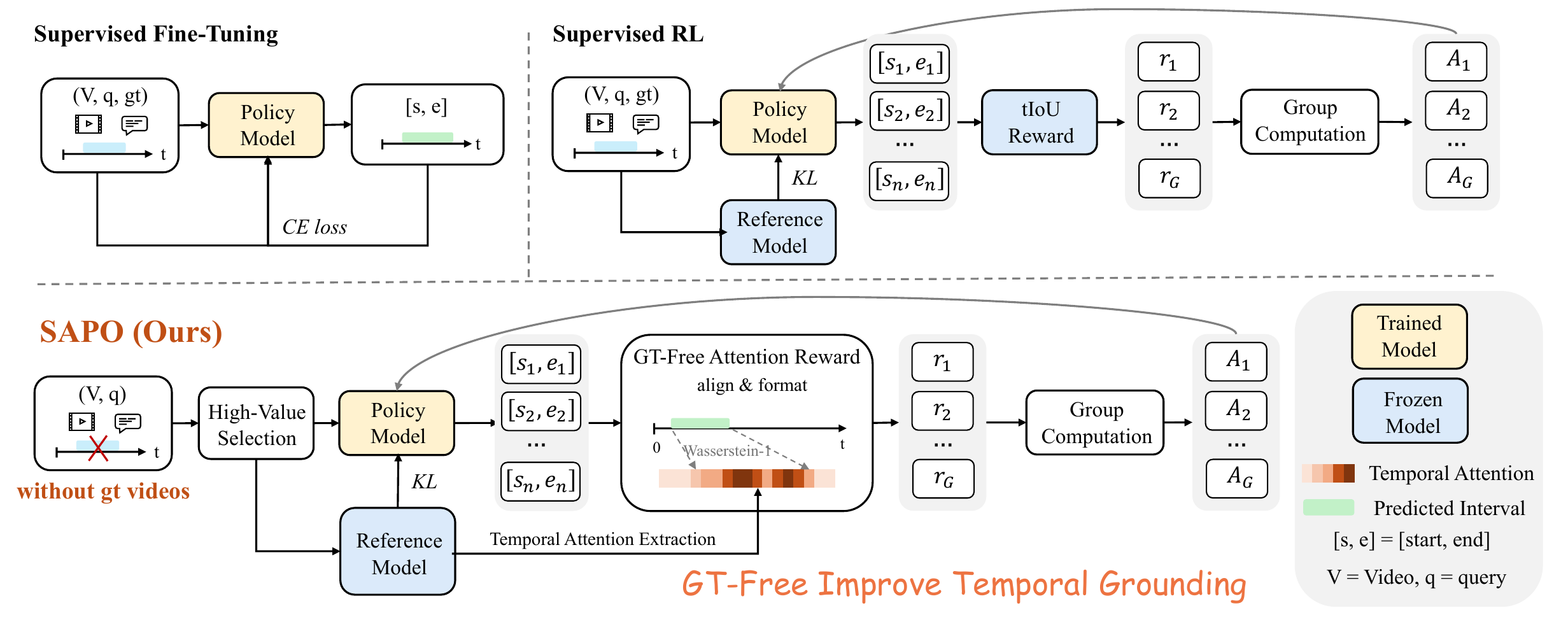}
    \caption{\textbf{Conventional TVG training and our annotation-free
    alternative.} \emph{Top:} Conventional TVG post-training relies on
    ground-truth temporal boundaries, using them either as supervised targets
    for fine-tuning or as rewards for reinforcement learning. \emph{Bottom:}
    SAPO instead derives self-supervision from the VLM's internal temporal
    attention to improve the model's temporal grounding ability without ground-truth boundaries.}
    \label{fig:framework}
\end{figure*}

Building on the preceding finding, we propose \textbf{Self-Supervised Attention
Policy Optimization (SAPO)}, which uses internal temporal attention as
self-supervision to improve the native grounding ability of a VLM. As
illustrated in Figure~\ref{fig:framework}, for each unlabeled video--query pair,
a trainable policy samples timestamp intervals, while a frozen reference model
provides a stationary temporal attention distribution. SAPO measures their
alignment as the rollout reward, estimates group-relative advantages, and
updates the policy. We next describe the complete optimization procedure and
the selection of high-value samples without ground truth. Implementation
details are provided in the Appendix.

\subsection{Self-Supervised Attention Policy Optimization}
Given an unlabeled sample $(V,q)$, the policy $\pi_\theta$ generates a group of
rollouts $\{\hat{y}_k\}_{k=1}^{K}$, each containing a timestamp interval
$I_{\hat{y}_k}=[s_k,e_k]$. In parallel, the reference model
$\pi_{\theta_0}$ produces a temporal attention profile over the video frames,
which is corrected and normalized into $a\in\Delta^T$. Decoding $a$ with
Eq.~\ref{eq:decode} gives the attention interval $I_a$. The frozen reference
model therefore provides two related forms of temporal supervision: the
continuous reference attention distribution $a$ and its decoded attention
interval $I_a$. The temporal distribution reward operates directly on $a$,
whereas the attention-interval IoU reward operates on $I_a$.

Directly using interval IoU as the reward produces sparse feedback: when
rollouts do not overlap with $I_a$, they all receive zero reward regardless of
their temporal distance from the attended region. SAPO therefore converts each
rollout interval into a continuous distribution $g(I_{\hat{y}})$ using a
Gaussian kernel and compares it with the corrected attention distribution $a$
before threshold-based interval decoding. SAPO scores each rollout by
\begin{equation}
\begin{aligned}
    r(\hat{y}) ={}& -w_{\mathrm{dist}} W_1\!\big(g(I_{\hat{y}}),a\big)
    + w_{\mathrm{IoU}}\,\mathrm{IoU}\!\big(I_{\hat{y}},I_a\big) \\
    &+ \lambda_{\mathrm{fmt}} r_{\mathrm{fmt}}
    - \lambda_{\mathrm{width}} p_{\mathrm{width}},
\end{aligned}
    \label{eq:reward}
\end{equation}

where $a$ provides attention-derived temporal supervision.
\textbf{(i) Temporal distribution reward.} The first term measures agreement
between the rollout-induced distribution and the reference attention
distribution using the Wasserstein-1 distance. Because $W_1$ increases with the temporal transport
required to match $a$, it provides graded feedback even for non-overlapping
rollouts.
\textbf{(ii) Attention-interval IoU reward.} The second term measures temporal
IoU between the generated interval and the interval $I_a$ decoded from the
reference attention distribution, providing a sparse interval-level complement to the
distribution reward.
\textbf{(iii) Format-validity reward.} $r_{\mathrm{fmt}}$ rewards a well-formed
interval ($0\!\le\!s\!<\!e\!\le\!\mathrm{dur}$), preventing malformed outputs.
\textbf{(iv) Interval-width regularization.} $p_{\mathrm{width}}$ penalizes
near-full-video intervals to discourage a trivial duration-prior solution.

For each video--query pair, we standardize the rewards within its rollout group
to obtain group-relative advantages and optimize the policy with GRPO
\cite{grpo2024}. The update increases the likelihood of intervals that align
more closely with the temporal attention while applying the standard KL
regularization. Repeating this process transfers the reference model's internal
temporal localization into the policy's explicit timestamp generation.

\subsection{High-Value Sample Selection}
The preceding objective can train on any unlabeled video--query pair, but not
every sample provides equally useful attention supervision. If the base
prediction already agrees with attention, there is little discrepancy for SAPO
to correct; if prediction is more accurate than attention, the internal signal
may not provide useful supervision. Determining which signal is more accurate
normally requires ground-truth boundaries. We therefore seek a proxy that does
not require ground truth and selects a higher proportion of samples on which
attention provides the
stronger localization signal. Given an unlabeled pool
$\mathcal{D}=\{(V_i,q_i)\}_{i=1}^{M}$, we run the base model once to obtain the
predicted interval $I_{p,i}$ and reference attention distribution $a_i$ for each
sample. We convert $I_{p,i}$ into a temporal distribution
$p_i=g(I_{p,i})$. Under a training budget of $k$ samples, we score each sample
by Jensen--Shannon divergence and select the top-$k$ examples:
\begin{equation}
    d_i=\operatorname{JS}(p_i\|a_i), \qquad
    \mathcal{D}_{\mathrm{sel}}
    =\operatorname{TopK}_{(V_i,q_i)\in\mathcal{D}}(d_i,k).
    \label{eq:selection}
\end{equation}

\begin{figure}[t]
    \centering
    \includegraphics[width=0.9\linewidth]{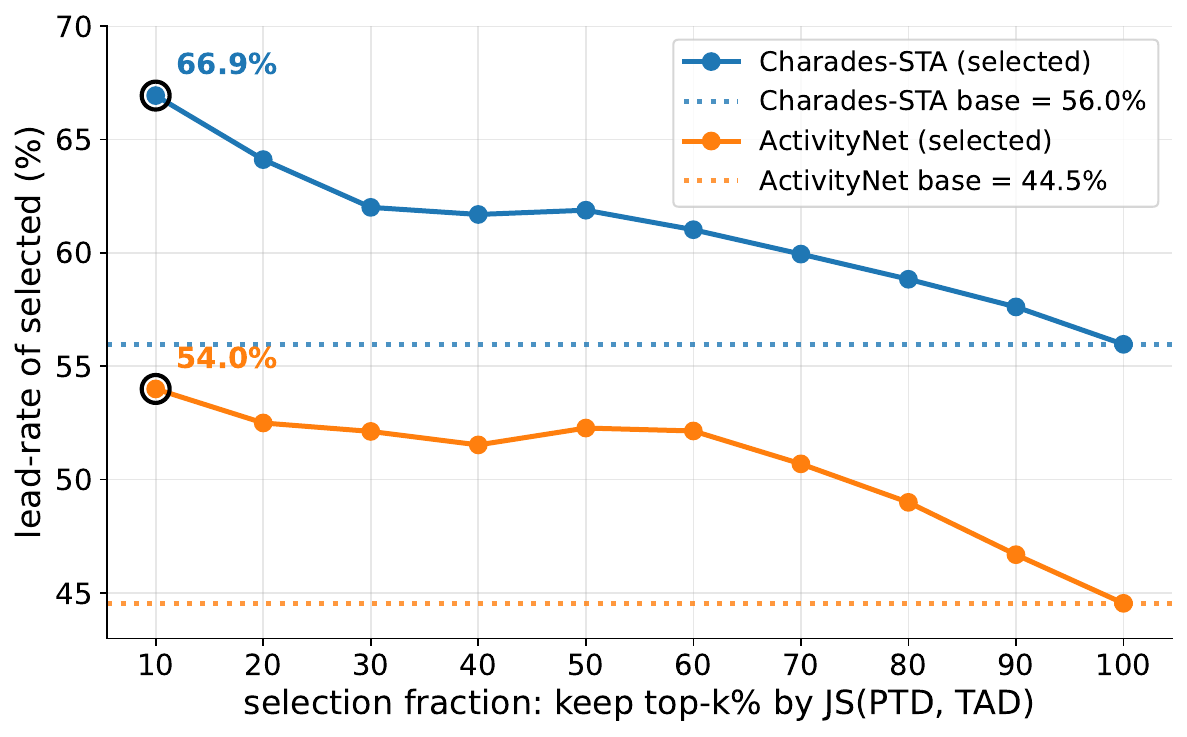}
    \caption{\textbf{Attention lead rate increases monotonically with
    attention--prediction JS divergence.} The rate is computed over all
    samples, with ties retained as non-leads.}
    \label{fig:topk_leadrate}
\end{figure}

As shown in Figure~\ref{fig:topk_leadrate}, the \emph{attention lead rate},
defined as the fraction of examples for which the attention interval is more
accurate than the predicted interval, increases with JS divergence on
both datasets. Unlike the conditional statistic in Table~\ref{tab:temporal_gap},
this rate is computed over all samples: ties are retained in the denominator
and counted as non-leads. The consistent trend indicates that
prediction--attention divergence can serve as a ground-truth-free proxy for high-value
samples. Because $d_i$ requires no temporal boundaries, the same top-$k$ rule
can be applied directly to a new unlabeled dataset. Further comparisons with
alternative selection criteria that do not use ground truth are reported in the experiments.

\noindent\textbf{Use of Ground Truth.}
Ground-truth boundaries are used only to validate the general diagnostic
findings and to evaluate final grounding
performance. Building on these findings, SAPO operates on training data with
temporal boundaries removed; its sample selection, reward construction, and
policy optimization require no ground truth.

\begin{table*}[t]
    \centering
    \small
    \setlength{\tabcolsep}{2.5pt}
    \begin{tabular}{lllcccccccccc}
        \toprule
        & & & \multicolumn{5}{c}{\textbf{Charades-STA}} & \multicolumn{5}{c}{\textbf{ActivityNet-Captions}} \\
        \cmidrule(lr){4-8}\cmidrule(lr){9-13}
        \textbf{Base} & \textbf{Method} & \textbf{train split} & R@0.3 & R@0.5 & R@0.7 & mIoU & $\Delta$ & R@0.3 & R@0.5 & R@0.7 & mIoU & $\Delta$ \\
        \midrule
        \multirow{4}{*}{Qwen2.5-VL-3B}
          & Zero-shot              & --                & 0.553 & 0.346 & 0.175 & 0.365 & -- & 0.342 & 0.216 & 0.119 & 0.246 & -- \\
          & \textbf{+ \method{}}   & Charades & 0.664 & 0.392 & 0.158 & 0.415 & \textbf{+13.7\%} & 0.409 & 0.241 & 0.126 & 0.281 & \textbf{+14.2\%} \\
          & \textbf{+ \method{}}   & ActivityNet & 0.629 & 0.368 & 0.165 & 0.398 & \textbf{+9.0\%} & 0.489 & 0.285 & 0.133 & 0.327 & \textbf{+32.9\%} \\
          & \textbf{+ \method{}}   & Charades+ActivityNet & 0.656 & 0.384 & 0.156 & 0.410 & \textbf{+12.3\%} & 0.493 & 0.287 & 0.130 & 0.329 & \textbf{+33.7\%} \\
        \midrule
        \multirow{4}{*}{Qwen2.5-VL-7B}
          & Zero-shot              & --                & 0.641 & 0.422 & 0.213 & 0.436 & -- & 0.381 & 0.238 & 0.130 & 0.274 & -- \\
          & \textbf{+ \method{}}   & Charades      & 0.671 & 0.443 & 0.218 & 0.459 & \textbf{+5.3\%} & 0.410 & 0.252 & 0.134 & 0.292 & \textbf{+6.6\%} \\
          & \textbf{+ \method{}}   & ActivityNet & 0.658 & 0.434 & 0.215 & 0.451 & \textbf{+3.4\%} & 0.522 & 0.303 & 0.142 & 0.348 & \textbf{+27.0\%} \\
          & \textbf{+ \method{}}   & \begin{tabular}[c]{@{}l@{}}Charades+ActivityNet\end{tabular} & 0.677 & 0.446 & 0.220 & 0.462 & \textbf{+6.0\%} & 0.501 & 0.293 & 0.137 & 0.334 & \textbf{+21.9\%} \\
        \midrule
        \multirow{4}{*}{Time-R1-7B}
          & Zero-shot              & --                & 0.815 & 0.578 & 0.270 & 0.525 & --     & 0.646 & 0.457 & 0.255 & 0.448 & -- \\
          & \textbf{+ \method{}}   & Charades     & 0.802 & 0.596 & 0.313 & 0.534 & \textbf{+1.7\%} & 0.657 & 0.470 & 0.263 & 0.456 & \textbf{+1.8\%} \\
          & \textbf{+ \method{}}   & ActivityNet & 0.805 & 0.588 & 0.301 & 0.530 & \textbf{+1.0\%} & 0.677 & 0.486 & 0.276 & 0.472 & \textbf{+5.4\%} \\
          & \textbf{+ \method{}}   & \begin{tabular}[c]{@{}l@{}}Charades+ActivityNet\end{tabular} & 0.782 & 0.610 & 0.335 & 0.533 & \textbf{+1.5\%} & 0.675 & 0.484 & 0.270 & 0.468 & \textbf{+4.5\%} \\
        \midrule
        \multicolumn{13}{l}{\textit{Training-Free Methods}} \\
          & ChatVTG & & 0.527 & 0.330 & 0.159 & 0.349 & -- & 0.407 & 0.225 & 0.094 & 0.272 & -- \\
          & TFVTG   & & 0.670 & 0.500 & 0.243 & 0.445 & -- & 0.493 & 0.270 & 0.134 & 0.341 & -- \\
          & TAG     & & 0.678 & 0.486 & 0.267 & 0.457 & -- & 0.519 & 0.289 & 0.151 & 0.366 & -- \\
        \bottomrule
    \end{tabular}
    \caption{\textbf{SAPO consistently improves temporal grounding across base
    models and training sources}.}
    \label{tab:main}
\end{table*}
\section{Experiments}

\subsection{Experimental Setup}
\noindent\textbf{Base Models.}
We apply the same SAPO procedure to three base models:
Qwen2.5-VL-3B and Qwen2.5-VL-7B \cite{qwen25vl2025}, which represent
general-purpose VLMs at two scales, and Time-R1-7B \cite{timer1}, a
state-of-the-art model with strong temporal grounding performance.

\noindent\textbf{Training Data.}
SAPO uses video--query pairs from the training splits of Charades-STA
\cite{charadessta2017} and ActivityNet-Captions
\cite{activitynetcaptions2017} after removing all temporal boundaries. Using
the ground-truth-free criterion described in the Method, we select 2,000
high-value samples from one dataset for single-source training or 1,000 from
each dataset for joint training, keeping the total budget fixed.

\noindent\textbf{Benchmarks and Metrics.}
We evaluate temporal grounding on the held-out test splits of Charades-STA and
ActivityNet-Captions under the standard train/test separation. To verify
that temporal adaptation does not substantially impair general video
understanding, we additionally report overall accuracy on Video-MME
\cite{videomme2024}. Full training configurations, implementation details, and
the Video-MME breakdown are provided in the Appendix.

\noindent\textbf{Baselines.}
Our primary baselines are the zero-shot versions of the three base models, which
enable controlled, paired comparisons under the same evaluation protocol. We
also report ChatVTG \cite{chatvtg}, TFVTG \cite{tfvtg2024}, and TAG
\cite{lee2025tag} as training-free numerical references because they share the
no-annotation condition. They are not strictly comparable, however, because
they improve inference through external workflows rather than update the
underlying model. To our knowledge, no existing TVG method provides a directly
matched setting that updates a model using video--query pairs without temporal
annotations. Unless otherwise specified, post-training analyses and ablations
use the joint-data setting.

\begin{table}[t]
    \centering
    \small
    \setlength{\tabcolsep}{7pt}
    \begin{tabular}{lcc}
        \toprule
        & \multicolumn{2}{c}{\textbf{Video-MME}} \\
        \cmidrule(lr){2-3}
        \textbf{Model} & Acc. & $\Delta$ \\
        \midrule
        Qwen2.5-VL-3B-Instruct                    & 58.0 & -- \\
        + \method{} trained on Charades           & 57.0 & -1.7\% \\
        + \method{} trained on ActivityNet        & 57.4 & -1.0\% \\
        \begin{tabular}[c]{@{}l@{}}+ \method{} trained on Charades+ActivityNet\end{tabular} & 57.4 & -1.0\% \\
        \midrule
        Time-R1-7B                                  & 60.1 & -- \\
        + \method{} trained on Charades             & 60.4 & +0.5\% \\
        + \method{} trained on ActivityNet          & 60.3 & +0.3\% \\
        + \method{} trained on Charades+ActivityNet & 60.3 & +0.3\% \\
        \bottomrule
    \end{tabular}
    \caption{\textbf{General video understanding on Video-MME}.}
    \label{tab:videomme}
\end{table}

\subsection{Main Results}
Table~\ref{tab:main} shows that SAPO consistently improves both general-purpose
VLMs and the already strong Time-R1 model. The gains are therefore not limited
to correcting a weak base model, but extend across model scales and initial
grounding abilities. Single-source training also transfers to the other
dataset, while joint training remains effective across evaluation domains.
These results indicate that SAPO learns a transferable improvement to native
timestamp generation rather than fitting a dataset-specific temporal prior,
supporting its use when adapting a VLM to previously unseen unlabeled data.
Published training-free results are included as numerical references, but are
not directly comparable because they improve inference through external
workflows rather than updating the model itself.

Table~\ref{tab:videomme} further shows that SAPO preserves general video
understanding while strengthening temporal grounding. Performance on Video-MME
remains close to that of the corresponding base model across training sources.

Because SAPO requires neither ground-truth boundaries nor supervised targets,
it also supports rapid adaptation directly on unlabeled target data. We
construct a joint target pool by selecting 1,000 examples from each benchmark's
test split, apply SAPO without accessing their temporal annotations, and then
evaluate the adapted model. Table~\ref{tab:settings} reports this explicitly
transductive protocol alongside adaptation on the corresponding joint training
pool. Target-set adaptation improves
the model beyond standard training-split adaptation, showing that SAPO can be
used in a practical test-time scenario: given a newly encountered unlabeled
target set, the model first adapts to its video--query pairs and is then
evaluated on that domain.
\begin{table}[t]
    \centering
    \small
    \setlength{\tabcolsep}{3pt}
    \begin{tabular}{lcccc}
        \toprule
        & \multicolumn{2}{c}{\textbf{Charades-STA}} &
          \multicolumn{2}{c}{\textbf{ActivityNet-Captions}} \\
        \cmidrule(lr){2-3}\cmidrule(lr){4-5}
        \textbf{Setting} & mIoU & $\Delta$ & mIoU & $\Delta$ \\
        \midrule
        Base (zero-shot)     & 0.365 & -- & 0.246 & -- \\
        SAPO from train sets & 0.410 & +12.3\% & 0.329 & +33.7\% \\
        SAPO from test sets  & 0.421 & +15.3\% & 0.341 & +38.6\% \\
        \bottomrule
    \end{tabular}
    \caption{\textbf{Standard and transductive adaptation with SAPO.}}
    \label{tab:settings}
\end{table}

\begin{table*}[t]
    \centering
    \small
    \setlength{\tabcolsep}{3pt}
    \begin{tabular}{lcccccccccc}
        \toprule
        & \multicolumn{4}{c}{\textbf{Predicted Interval: Charades-STA}} &
          \multicolumn{4}{c}{\textbf{Predicted Interval: ActivityNet}} &
          \multicolumn{2}{c}{\textbf{Attention Interval}} \\
        \cmidrule(lr){2-5}\cmidrule(lr){6-9}\cmidrule(lr){10-11}
        \textbf{Variant}
        & \textbf{R@0.3} & \textbf{R@0.5} & \textbf{R@0.7} & \textbf{mIoU}
        & \textbf{R@0.3} & \textbf{R@0.5} & \textbf{R@0.7} & \textbf{mIoU}
        & \textbf{Char.\ mIoU} & \textbf{Act.\ mIoU} \\
        \midrule
        Base
            & 0.553 & 0.346 & \textbf{0.175} & 0.365
            & 0.342 & 0.216 & 0.119 & 0.246
            & \underline{0.395} & \underline{0.328} \\
        \midrule
        Attention pseudo-label + SFT
            & 0.647 & \underline{0.371} & 0.149 & 0.395
            & 0.480 & 0.247 & 0.102 & 0.315
            & 0.394 & 0.322 \\
        \midrule
        \textbf{SAPO (Full)}
            & 0.656 & \textbf{0.384} & 0.156 & \textbf{0.410}
            & 0.493 & \underline{0.287} & \textbf{0.130} & \textbf{0.329}
            & \textbf{0.401} & \textbf{0.335} \\
        \enspace 1 w/o attention-derived rewards
            & 0.492 & 0.312 & 0.159 & 0.330
            & 0.226 & 0.113 & 0.054 & 0.163
            & -- & -- \\
        \enspace 2 w/o temporal distribution reward
            & 0.597 & 0.362 & \underline{0.161} & 0.378
            & 0.406 & 0.249 & \underline{0.122} & 0.277
            & -- & -- \\
        \enspace 3 w/o attention-interval IoU reward
            & 0.616 & 0.352 & 0.145 & 0.388
            & 0.471 & \textbf{0.295} & 0.114 & 0.312
            & -- & -- \\
        \enspace 4 replace $W_1$ with JS divergence
            & \underline{0.673} & 0.362 & 0.141 & \underline{0.404}
            & \underline{0.532} & 0.272 & 0.119 & 0.323
            & -- & -- \\
        \enspace 5 w/o interval-width regularization
            & \textbf{0.687} & 0.335 & 0.129 & 0.403
            & \textbf{0.534} & 0.276 & 0.116 & \underline{0.325}
            & -- & -- \\
 
        \bottomrule
    \end{tabular}
    \caption{\textbf{Ablation of SAPO}.
    The first eight columns evaluate predicted intervals, while the last two
    report the mIoU of attention intervals. Variants 1--5 are discussed in the
    same order in the ablation analysis. Best and second-best results are marked
    in bold and underlined, respectively.}
    \label{tab:ablation_components}
\end{table*}
\subsection{In-Depth Analysis}
We examine whether SAPO transfers internal localization evidence into native
timestamps, closes the temporal generation gap, and changes attention accuracy,
while also comparing it with direct attention decoding.

\begin{figure}[t]
    \centering
    \includegraphics[width=\columnwidth]{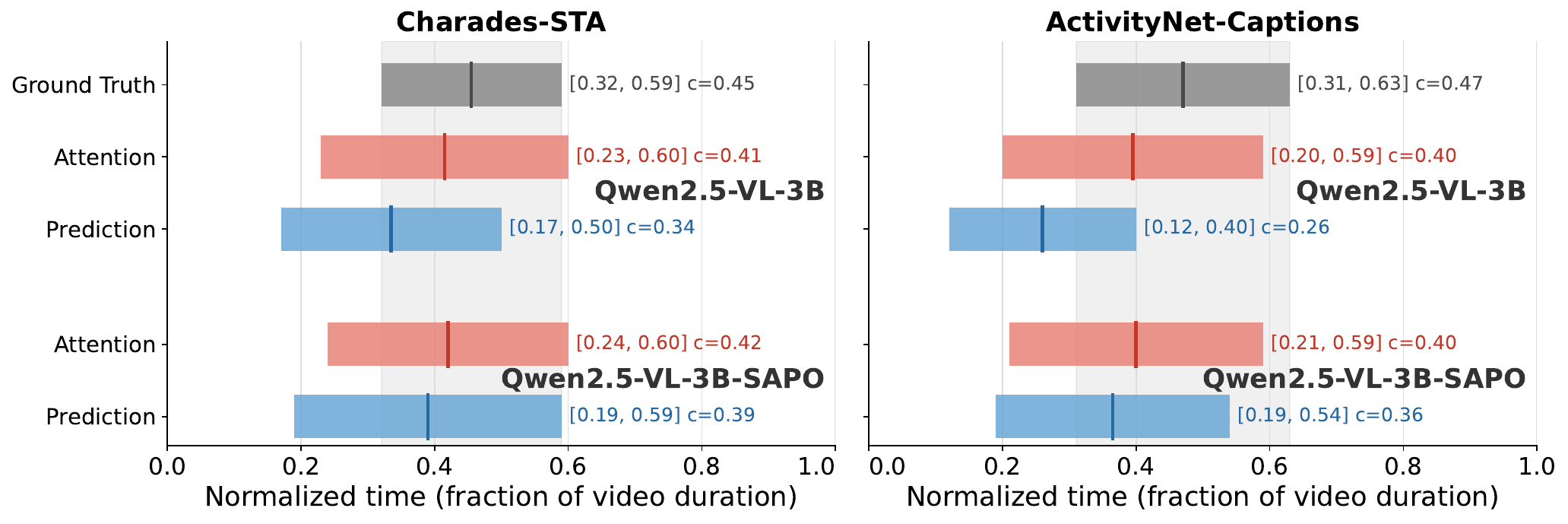}
    \caption{\textbf{SAPO moves the average predicted interval toward the
    ground-truth temporal region.}}
    \label{fig:avg_intervals_before_after}
\end{figure}

\textbf{Does SAPO close the temporal generation gap?}
As shown in Table~\ref{tab:ablation_components}, SAPO substantially improves native
timestamp prediction on both datasets and reduces the prediction--attention
gap to within one mIoU point. The trained prediction slightly exceeds attention
on Charades-STA and nearly matches it on ActivityNet-Captions, indicating that
the original generation gap is effectively closed.
Figure~\ref{fig:avg_intervals_before_after} provides complementary visual
evidence: compared with the base model, the SAPO prediction aligns more closely
with both the attention interval and the ground-truth event region.

Attention accuracy remains largely stable after SAPO, with only minor increases
on both datasets that may arise because prediction and attention share model
parameters; we therefore treat this change as a secondary effect.

\begin{figure}[t]
    \centering
    \includegraphics[width=\columnwidth]{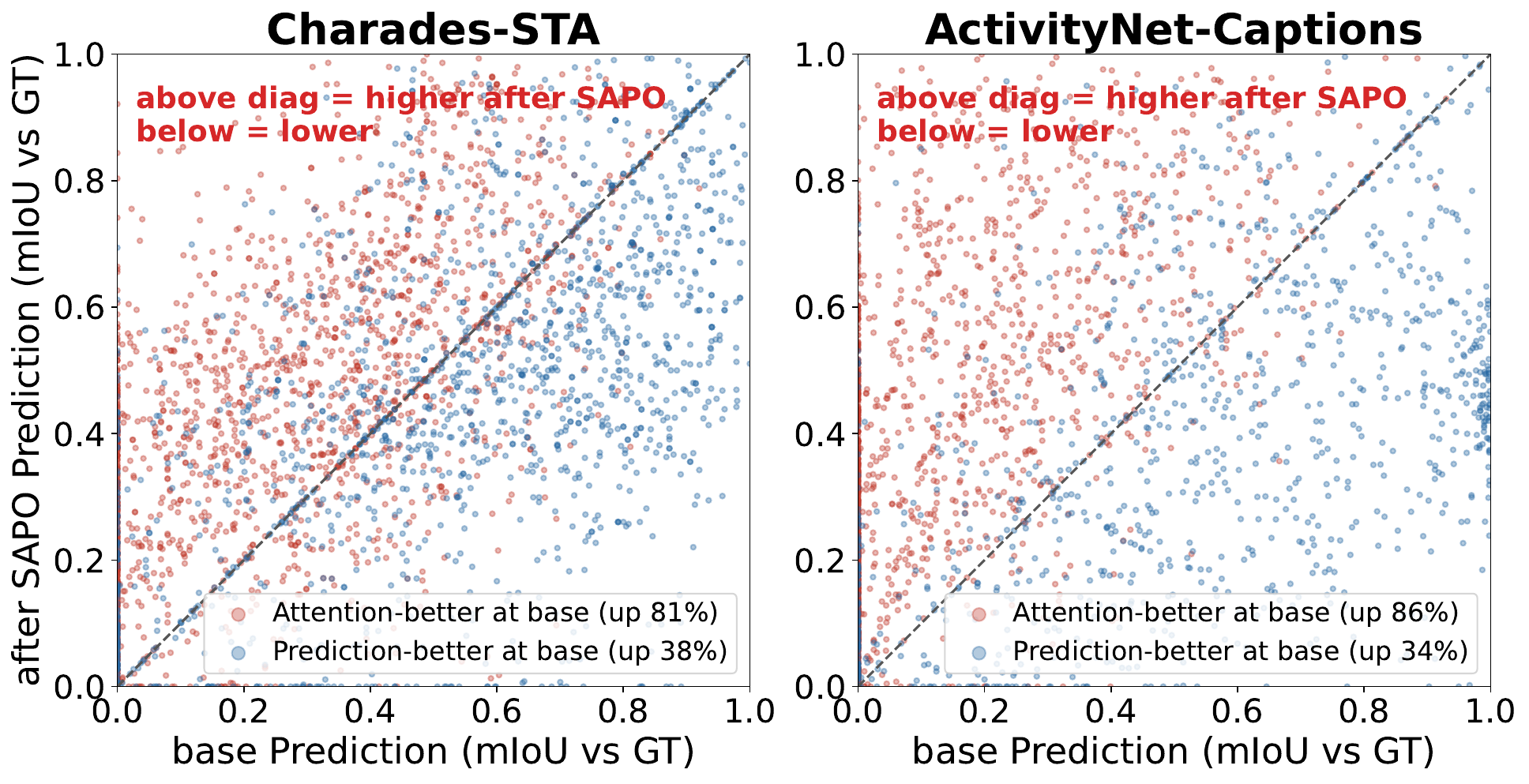}
    \caption{\textbf{SAPO preferentially improves samples whose base attention
    is more accurate than their base prediction.}}
    \label{fig:why_surpass_teacher}
\end{figure}

\textbf{How does SAPO compare with direct attention decoding?}
As shown in Table~\ref{tab:ablation_components}, the native SAPO prediction achieves
higher aggregate accuracy than directly decoding its attention. To
explain this seemingly counter-intuitive result, we partition queries using the
initial model: frozen attention has higher ground-truth IoU in the
\emph{attention-better} group, while the remaining queries form the
\emph{prediction-better} group. Each point in
Figure~\ref{fig:why_surpass_teacher} compares one query's initial and post-SAPO
prediction IoUs, with the diagonal denoting no change. SAPO improves $81\%$/$86\%$ of
attention-better queries and, despite optimizing toward attention, also improves
$38\%$/$34\%$ of prediction-better queries on Charades-STA/ActivityNet-Captions.
Thus, gains on most attention-better queries outweigh regressions while many
prediction-better cases are retained or further improved. Ground-truth-free selection
further concentrates training on samples where attention is the stronger
reference signal, allowing native predictions to exceed it in aggregate without doing so
on every query.

SAPO also retains native timestamp generation at inference, unlike direct
decoding, which requires white-box attention access, layer selection,
correction, and temporal-window extraction for every test example.

\section{Ablation Studies}
We ablate two main design choices: the selection of high-value samples without
ground truth and the reward for policy optimization. All experiments use the
Qwen2.5-VL-3B base model and the joint training pool from Charades and
ActivityNet.

\noindent\textbf{Training Sample Selector.}
We first assess how top-$k$ selection based on prediction--attention JS
divergence affects downstream grounding performance. As shown in
Table~\ref{tab:ablation_select}, random selection already improves the base
model because the original unlabeled pool contains many attention-better
examples. Under the same sample budget, JS selection yields consistently
stronger downstream performance by increasing the proportion of such
high-value examples. We also report a ground-truth oracle that selects only
attention-better samples as an upper-bound reference; this diagnostic upper
bound is not part of standard SAPO, which never accesses ground truth.

\begin{figure}[t]
    \centering
    \includegraphics[width=0.95\linewidth]{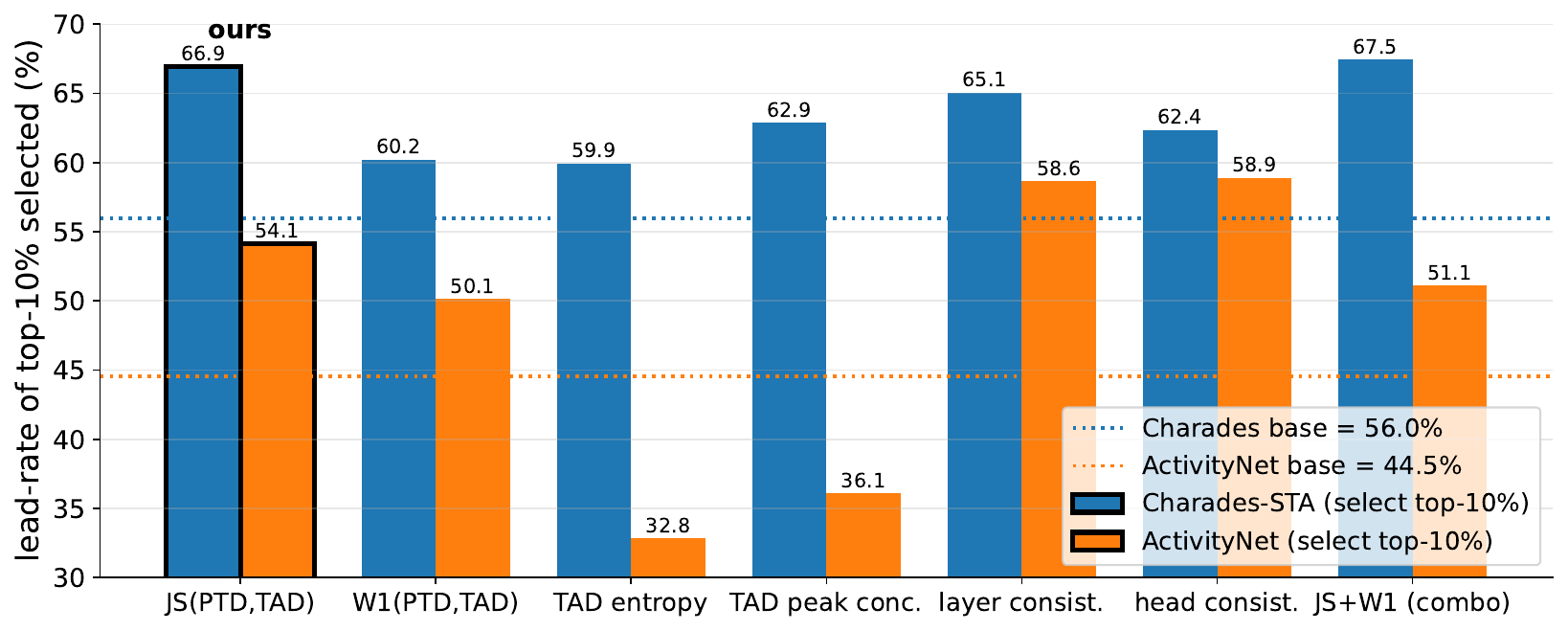}
    \caption{\textbf{Comparison of selection metrics.}}
    \label{fig:metric_ablation}
\end{figure}

\begin{table}[t]
    \centering
    \footnotesize
    \setlength{\tabcolsep}{1.2pt}

    \begin{tabular}{lcccccc}
        \toprule
        & \multicolumn{3}{c}{\textbf{Charades-STA}} & \multicolumn{3}{c}{\textbf{ActivityNet-Captions}} \\
        \cmidrule(lr){2-4}\cmidrule(lr){5-7}
        \textbf{Selector} & \textbf{R@0.5} & \textbf{mIoU} & $\boldsymbol{\Delta}$
        & \textbf{R@0.5} & \textbf{mIoU} & $\boldsymbol{\Delta}$ \\
        \midrule
        Base (no training)      & 0.346 & 0.365 & -- & 0.216 & 0.246 & -- \\
        \midrule
        \multicolumn{7}{l}{\textit{SAPO with different training-sample selectors}} \\
        \quad Random
            & 0.369 & 0.396 & +8.5\% & 0.271 & 0.312 & +26.8\% \\
        \quad \textbf{JS divergence}
            & \textbf{0.384} & \textbf{0.410} & \textbf{+12.3\%} & \textbf{0.287} & \textbf{0.329} & \textbf{+33.7\%} \\
        \quad \textit{GT oracle}$^\dagger$
            & \textit{0.386} & \textit{0.423} & \textit{+15.9\%} & \textit{0.292} & \textit{0.343} & \textit{+39.4\%} \\
        \bottomrule
    \end{tabular}
    \caption{\textbf{Downstream TVG performance under different training-set
    selection strategies}. The variants differ only in the selected training
    samples; $\Delta$ denotes the relative
    mIoU improvement over the base model.}
    \label{tab:ablation_select}
\end{table}
\noindent\textbf{Selection Metric.}
Beyond the random comparison, we evaluate alternative ranking metrics that do
not use ground truth under the same selection budget.
Figure~\ref{fig:metric_ablation} shows that prediction--attention JS divergence
produces a higher attention lead rate, indicating that it retrieves a sample
set with a larger proportion of high-value examples. Together with the
downstream gains in Table~\ref{tab:ablation_select}, this result supports JS
divergence as an effective criterion for constructing informative RL data
without ground truth.

\noindent\textbf{Reward Components.}
Table~\ref{tab:ablation_components} compares Full SAPO with five controlled
variants that modify one aspect of the reward while keeping the training data
and optimization settings fixed. We report mIoU together with recall at
multiple IoU thresholds to assess boundary precision more comprehensively. The
final two columns additionally track
whether training changes the attention interval itself. Variant 1 removes both
attention-derived rewards and performs below the base model, confirming that
the gains of SAPO arise from attention-based self-supervision rather than
generic format or width constraints.

SAPO uses attention supervision at two complementary levels. The temporal
distribution reward aligns each rollout distribution with the full reference
attention distribution through $W_1$, providing dense feedback that reflects
temporal distance. The attention-interval IoU reward instead aligns the
generated interval with the decoded attention interval, providing sparse
boundary-level supervision. Variants 2 and 3 show that both components are
effective. The larger degradation in Variant 2 further establishes the
importance of dense supervision: interval IoU cannot distinguish
non-overlapping rollouts, whereas distribution alignment still reflects their
distances from the attended region.
Variant 4 replaces $W_1$ with JS. Although the two variants achieve similar
mIoU, JS reduces recall at stricter IoU thresholds. Unlike JS, $W_1$ preserves
temporal geometry and provides distance-sensitive feedback for separated
distributions, leading to more precise boundaries.
Variant 5 exposes the role of width regularization. Without it, the model
expands its predicted intervals to obtain coarse overlap, keeping mIoU similar
while substantially degrading high-precision localization. The pseudo-label
SFT baseline is instead constrained by the accuracy of its hard attention
targets: its predicted intervals approach but remain below Full SAPO, and its
attention accuracy remains essentially unchanged. In contrast, SAPO improves
both predicted and attention intervals, indicating a joint adaptation mechanism
beyond direct imitation of hard attention targets.

\section{Conclusion}
We identify a temporal generation gap in video VLMs: attention intervals
localize queried events more accurately than predicted intervals on
approximately $66\%$ of valid examples. This finding reveals internal temporal
evidence that can supervise timestamp generation without human-annotated
boundaries. Based on this insight, we propose \method{}, which uses temporal
attention as self-supervision for group-relative policy optimization to improve
native timestamp prediction. Across multiple models, datasets, and training sources,
\method{} improves temporal grounding for both general-purpose VLMs and a
strong temporal grounding model, transfers across datasets, and largely
preserves general video understanding.

\bibliography{sapo}

@article{qwen25vl2025,
  title   = {Qwen2.5-VL Technical Report},
  author  = {Bai, Shuai and Chen, Keqin and Liu, Xuejing and Wang, Jialin and others},
  journal = {arXiv preprint arXiv:2502.13923},
  year    = {2025}
}

@inproceedings{timelens2026,
  title     = {{TimeLens}: Rethinking Video Temporal Grounding with Multimodal {LLMs}},
  author    = {Zhang, Jun and Wang, Teng and Ge, Yuying and Ge, Yixiao and Li, Xinhao and Shan, Ying and Wang, Limin},
  booktitle = {Proceedings of the IEEE/CVF Conference on Computer Vision and Pattern Recognition (CVPR)},
  pages     = {10419--10429},
  year      = {2026}
}

@article{internvl25,
  title   = {Expanding Performance Boundaries of Open-Source Multimodal Models with Model, Data, and Test-Time Scaling},
  author  = {Chen, Zhe and Wang, Weiyun and Cao, Yue and others},
  journal = {arXiv preprint arXiv:2412.05271},
  year    = {2024}
}

@article{llavaonevision,
  title         = {{LLaVA-OneVision}: Easy Visual Task Transfer},
  author        = {Li, Bo and Zhang, Yuanhan and Guo, Dong and Zhang, Renrui and Li, Feng and Zhang, Hao and Zhang, Kaichen and Zhang, Peiyuan and Li, Yanwei and Liu, Ziwei and Li, Chunyuan},
  journal       = {arXiv preprint arXiv:2408.03326},
  year          = {2024},
  eprint        = {2408.03326},
  archivePrefix = {arXiv},
  primaryClass  = {cs.CV}
}

@article{grpo2024,
  title   = {DeepSeekMath: Pushing the Limits of Mathematical Reasoning in Open Language Models},
  author  = {Shao, Zhihong and Wang, Peiyi and Zhu, Qihao and others},
  journal = {arXiv preprint arXiv:2402.03300},
  year    = {2024}
}

@inproceedings{timer1,
  title     = {{Time-R1}: Post-Training Large Vision Language Model for Temporal Video Grounding},
  author    = {Wang, Ye and Wang, Ziheng and Xu, Boshen and Du, Yang and Lin, Kejun and Xiao, Zihan and Yue, Zihao and Ju, Jianzhong and Zhang, Liang and Yang, Dingyi and Fang, Xiangnan and He, Zewen and Luo, Zhenbo and Wang, Wenxuan and Lin, Junqi and Luan, Jian and Jin, Qin},
  booktitle = {Advances in Neural Information Processing Systems},
  year      = {2025},
  url       = {https://openreview.net/forum?id=gJ05Gm5VxQ}
}

@inproceedings{vtimellm,
  title     = {VTimeLLM: Empower LLM to Grasp Video Moments},
  author    = {Huang, Bin and Wang, Xin and Chen, Hong and Song, Zihan and Zhu, Wenwu},
  booktitle = {Proceedings of the IEEE/CVF Conference on Computer Vision and Pattern Recognition (CVPR)},
  year      = {2024}
}

@inproceedings{timechat,
  title     = {TimeChat: A Time-Sensitive Multimodal Large Language Model for Long Video Understanding},
  author    = {Ren, Shuhuai and Yao, Linli and Li, Shicheng and Sun, Xu and Hou, Lu},
  booktitle = {Proceedings of the IEEE/CVF Conference on Computer Vision and Pattern Recognition (CVPR)},
  year      = {2024}
}

@article{hawkeye,
  title   = {{HawkEye}: Training Video-Text {LLMs} for Grounding Text in Videos},
  author  = {Wang, Yueqian and Meng, Xiaojun and Liang, Jianxin and Wang, Yuxuan and Liu, Qun and Zhao, Dongyan},
  journal = {arXiv preprint arXiv:2403.10228},
  year    = {2024}
}

@inproceedings{trace,
  title     = {{TRACE}: Temporal Grounding Video {LLM} via Causal Event Modeling},
  author    = {Guo, Yongxin and Liu, Jingyu and Li, Mingda and Liu, Qingbin and Chen, Xi and Tang, Xiaoying},
  booktitle = {Proceedings of the International Conference on Learning Representations (ICLR)},
  year      = {2025}
}

@inproceedings{timesuite,
  title     = {{TimeSuite}: Improving {MLLMs} for Long Video Understanding via Grounded Tuning},
  author    = {Zeng, Xiangyu and Li, Kunchang and Wang, Chenting and Li, Xinhao and Jiang, Tianxiang and Yan, Ziang and Li, Songze and Shi, Yansong and Yue, Zhengrong and Wang, Yi and Wang, Yali and Qiao, Yu and Wang, Limin},
  booktitle = {Proceedings of the International Conference on Learning Representations (ICLR)},
  year      = {2025}
}

@inproceedings{groundedvideollm,
  title     = {Grounded-{VideoLLM}: Sharpening Fine-grained Temporal Grounding in Video Large Language Models},
  author    = {Wang, Haibo and Xu, Zhiyang and Cheng, Yu and Diao, Shizhe and Zhou, Yufan and Cao, Yixin and Wang, Qifan and Ge, Weifeng and Huang, Lifu},
  booktitle = {Findings of the Association for Computational Linguistics: EMNLP},
  year      = {2025}
}

@inproceedings{vtgllm,
  title     = {{VTG-LLM}: Integrating Timestamp Knowledge into Video {LLMs} for Enhanced Video Temporal Grounding},
  author    = {Guo, Yongxin and Liu, Jingyu and Li, Mingda and Cheng, Dingxin and Tang, Xiaoying and Sui, Dianbo and Liu, Qingbin and Chen, Xi and Zhao, Kevin},
  booktitle = {Proceedings of the AAAI Conference on Artificial Intelligence},
  year      = {2025}
}

@inproceedings{numpro,
  title     = {Number it: Temporal Grounding Videos like Flipping Manga},
  author    = {Wu, Yongliang and Hu, Xinting and Sun, Yuyang and Zhou, Yizhou and Zhu, Wenbo and Rao, Fengyun and Schiele, Bernt and Yang, Xu},
  booktitle = {Proceedings of the IEEE/CVF Conference on Computer Vision and Pattern Recognition (CVPR)},
  year      = {2025}
}

@article{videoopd,
  title   = {Video-{OPD}: Efficient Post-Training of Multimodal Large Language Models for Temporal Video Grounding via On-Policy Distillation},
  author  = {Li, Jiaze and Yin, Hao and Xu, Haoran and Xu, Boshen and Tan, Wenhui and He, Zewen and Ju, Jianzhong and Luo, Zhenbo and Luan, Jian},
  journal = {arXiv preprint arXiv:2602.02994},
  year    = {2026}
}

@inproceedings{revisionllm,
  title     = {{ReVisionLLM}: Recursive Vision-Language Model for Temporal Grounding in Hour-Long Videos},
  author    = {Hannan, Tanveer and Islam, Md Mohaiminul and Gu, Jindong and Seidl, Thomas and Bertasius, Gedas},
  booktitle = {Proceedings of the IEEE/CVF Conference on Computer Vision and Pattern Recognition (CVPR)},
  pages     = {19012--19022},
  year      = {2025}
}

@inproceedings{lee2025tag,
  title     = {{TAG}: A Simple Yet Effective Temporal-Aware Approach for Zero-Shot Video Temporal Grounding},
  author    = {Lee, Jin-Seop and Lee, SungJoon and Ahn, Jaehan
               and Choi, YunSeok and Lee, Jee-Hyong},
  booktitle = {Proceedings of the British Machine Vision Conference},
  year      = {2025}
}

@inproceedings{tfvtg2024,
  title     = {Training-Free Video Temporal Grounding Using
               Large-Scale Pre-Trained Models},
  author    = {Zheng, Minghang and Cai, Xinhao and Chen, Qingchao
               and Peng, Yuxin and Liu, Yang},
  booktitle = {Proceedings of the European Conference on Computer Vision},
  pages     = {20--37},
  year      = {2024},
  doi       = {10.1007/978-3-031-73007-8_2}
}

@inproceedings{chatvtg,
  title     = {{ChatVTG}: Video Temporal Grounding via Chat with Video Dialogue Large Language Models},
  author    = {Qu, Mengxue and Chen, Xiaodong and Liu, Wu and Li, Alicia and Zhao, Yao},
  booktitle = {Proceedings of the IEEE/CVF Conference on Computer Vision and Pattern Recognition Workshops (CVPRW)},
  pages     = {1847--1856},
  year      = {2024}
}

@inproceedings{2dtan,
  title     = {Learning 2D Temporal Adjacent Networks for Moment Localization with Natural Language},
  author    = {Zhang, Songyang and Peng, Houwen and Fu, Jianlong and Luo, Jiebo},
  booktitle = {Proceedings of the AAAI Conference on Artificial Intelligence},
  year      = {2020}
}

@inproceedings{momentdetr,
  title     = {Detecting Moments and Highlights in Videos via Natural Language Queries},
  author    = {Lei, Jie and Berg, Tamara L. and Bansal, Mohit},
  booktitle = {Advances in Neural Information Processing Systems (NeurIPS)},
  year      = {2021}
}

@inproceedings{charadessta2017,
  title     = {TALL: Temporal Activity Localization via Language Query},
  author    = {Gao, Jiyang and Sun, Chen and Yang, Zhenheng and Nevatia, Ram},
  booktitle = {Proceedings of the IEEE International Conference on Computer Vision (ICCV)},
  year      = {2017}
}

@inproceedings{activitynetcaptions2017,
  title     = {Dense-Captioning Events in Videos},
  author    = {Krishna, Ranjay and Hata, Kenji and Ren, Frederic and Fei-Fei, Li and Niebles, Juan Carlos},
  booktitle = {Proceedings of the IEEE International Conference on Computer Vision (ICCV)},
  year      = {2017}
}

@article{videomme2024,
  title   = {Video-MME: The First-Ever Comprehensive Evaluation Benchmark of Multi-modal LLMs in Video Analysis},
  author  = {Fu, Chaoyou and Dai, Yuhan and Luo, Yondong and others},
  journal = {arXiv preprint arXiv:2405.21075},
  year    = {2024}
}

@article{zhao2024first,
  title   = {The First to Know: How Token Distributions Reveal Hidden Knowledge in Large Vision-Language Models?},
  author  = {Zhao, Qinyu and Xu, Ming and Gupta, Kartik and Asthana, Akshay and Zheng, Liang and Gould, Stephen},
  journal = {arXiv preprint arXiv:2403.09037},
  year    = {2024}
}

@inproceedings{Palit_2023_ICCV,
  title     = {Towards Vision-Language Mechanistic Interpretability: A Causal Tracing Tool for {BLIP}},
  author    = {Palit, Vedant and Pandey, Rohan and Arora, Aryaman and Liang, Paul Pu},
  booktitle = {Proceedings of the IEEE/CVF International Conference on Computer Vision Workshops},
  pages     = {2856--2861},
  year      = {2023}
}

\appendix
\clearpage
\section{Appendix}

\subsection{Attention Extraction Protocol}
\label{sec:attention_extraction}
\noindent\textbf{Evaluation data.}
The temporal generation gap in Table~\ref{tab:temporal_gap} is measured on all
3,720 Charades-STA examples and a fixed random sample of 4,000 examples from the
17,505 ActivityNet examples. The identical subsets are used for all four VLMs.
Ground-truth intervals are accessed only to compute the diagnostic IoU values;
they are not used to extract attention, choose examples, or train standard
SAPO. The explicitly marked GT-oracle ablation is reported only as an upper
bound.

\begin{figure}[t]
    \centering
    \begin{minipage}{0.48\linewidth}
        \centering
        \includegraphics[width=\linewidth]{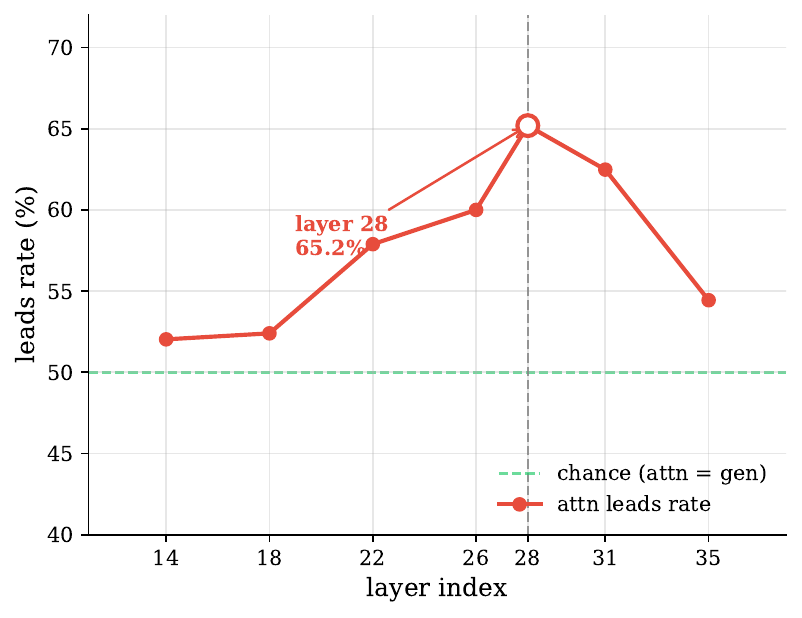}
        \\[-1pt]{\small (a) Layer sensitivity (top-$p=0.5$)}
    \end{minipage}
    \hfill
    \begin{minipage}{0.48\linewidth}
        \centering
        \includegraphics[width=\linewidth]{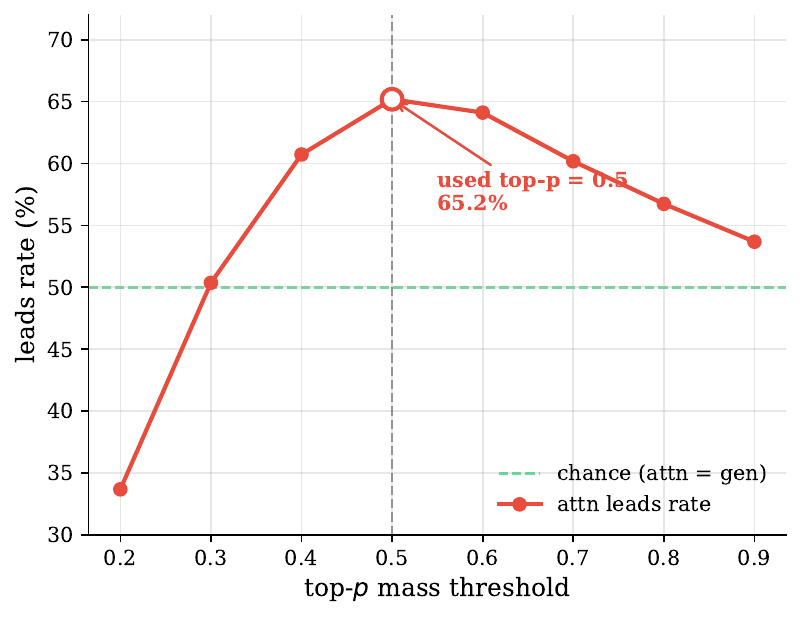}
        \\[-1pt]{\small (b) Threshold sensitivity (layer L28)}
    \end{minipage}
    \caption{\textbf{Diagnostic sensitivity analysis} for Qwen2.5-VL-3B on
    Charades-STA, shown as a representative example. We report the
    \emph{attention lead rate}: the fraction of examples on which the
    attention interval attains a higher IoU with the ground truth than the
    model's own predicted interval, excluding ties. \textbf{(a)} The lead rate
    remains stable across the examined middle-to-late layers.
    \textbf{(b)} The same conclusion holds across a range of attention-mass
    thresholds around the fixed value $0.5$.}
    \label{fig:attn_sensitivity}
\end{figure}

\noindent\textbf{Diagnostic attention extraction.}
We take the last prompt token's cross-modal attention to video tokens from a
fixed mid-to-late layer, average it over attention heads, and pool the spatial
visual tokens belonging to each frame. After removing the first-bin attention
sink, we renormalize the resulting temporal profile. Its attention interval is
the shortest contiguous window covering $50\%$ of the cumulative attention
mass, as defined in Eq.~\ref{eq:decode}, and is then mapped from frame positions
to seconds. We use ground-truth boundaries only in this offline mechanism
analysis to measure diagnostic IoU. Table~\ref{tab:attn_layers} lists the layer
used for each model; the same layer is used on both datasets and throughout all
experiments. Figure~\ref{fig:attn_sensitivity} reports sensitivity across
middle-to-late layers and attention-mass thresholds.

\noindent\textbf{Reference attention distribution.}
The reference attention distribution used during training follows the same last-token
extraction, head averaging, spatial pooling, and interval-decoding procedure.
We additionally divide the temporal profile by a position prior estimated from
the training pool without temporal boundaries to remove dataset-level temporal
bias, and
renormalize the result into a distribution $a\in\Delta^{T}$. The attention
interval $I_a$ is the shortest contiguous window covering $50\%$ of the
corrected attention mass. The layer and threshold remain fixed throughout
sample selection and SAPO adaptation. For each model architecture, we directly
specify one representative middle-to-late layer without ranking candidate
layers by ground-truth grounding accuracy. The same layer is then used across
datasets; the sensitivity analysis below verifies that the diagnostic
conclusion is stable across nearby middle-to-late layers.

\begin{table}[h]
    \centering
    \small
    \setlength{\tabcolsep}{2pt}
    \begin{tabular}{llc}
        \toprule
        \textbf{Model} & \textbf{Dataset} & \textbf{Layer} \\
        \midrule
        Qwen2.5-VL-3B & Charades-STA & L28 \\
        Qwen2.5-VL-3B & ActivityNet  & L28 \\
        Qwen2.5-VL-7B & Charades-STA & L22 \\
        Qwen2.5-VL-7B & ActivityNet  & L22 \\
        Qwen3-VL-4B   & Charades-STA & L24 \\
        Qwen3-VL-4B   & ActivityNet  & L24 \\
        InternVL3-2B  & Charades-STA & L26 \\
        InternVL3-2B  & ActivityNet  & L26 \\
        \bottomrule
    \end{tabular}
    \caption{\textbf{Attention layers used for temporal interval decoding.}
    A representative middle-to-late layer is specified without ground-truth
    layer selection and kept fixed across both datasets and all experiments.}
    \label{tab:attn_layers}
\end{table}

\subsection{Detailed Generation-Gap Analysis}
\noindent\textbf{Pre-SAPO group-wise attention--prediction analysis.}
To complement the aggregate results in Table~\ref{tab:temporal_gap}, we
partition each model--dataset pair according to whether temporal attention
decoding (TAD) or predicted timestamp decoding (PTD) attains higher
ground-truth IoU. Table~\ref{tab:base_group_miou} reports the TAD and PTD mIoU
within the attention-better, prediction-better, and all-zero tie groups.

\begin{table*}[t]
    \centering
    \small
    \setlength{\tabcolsep}{4pt}
    \begin{tabular}{llcccccc}
        \toprule
        & & \multicolumn{2}{c}{\textbf{Attention Better}} &
        \multicolumn{2}{c}{\textbf{Prediction Better}} &
        \multicolumn{2}{c}{\textbf{Tie (Both Zero)}} \\
        \cmidrule(lr){3-4}\cmidrule(lr){5-6}\cmidrule(lr){7-8}
        \textbf{Model} & \textbf{Dataset}
        & \textbf{TAD} & \textbf{PTD}
        & \textbf{TAD} & \textbf{PTD}
        & \textbf{TAD} & \textbf{PTD} \\
        \midrule
        Qwen2.5-VL-3B & Charades-STA
            & 0.515 & 0.268 & 0.298 & 0.663 & 0.000 & 0.000 \\
        InternVL3-2B & Charades-STA
            & 0.489 & 0.077 & 0.227 & 0.511 & 0.000 & 0.000 \\
        Qwen2.5-VL-7B & Charades-STA
            & 0.587 & 0.346 & 0.368 & 0.689 & 0.000 & 0.000 \\
        Qwen3-VL-4B & Charades-STA
            & 0.617 & 0.359 & 0.360 & 0.660 & 0.000 & 0.000 \\
        \midrule
        Qwen2.5-VL-3B & ActivityNet-Captions
            & 0.412 & 0.124 & 0.270 & 0.656 & 0.000 & 0.000 \\
        InternVL3-2B & ActivityNet-Captions
            & 0.374 & 0.087 & 0.171 & 0.437 & 0.000 & 0.000 \\
        Qwen2.5-VL-7B & ActivityNet-Captions
            & 0.418 & 0.138 & 0.242 & 0.642 & 0.000 & 0.000 \\
        Qwen3-VL-4B & ActivityNet-Captions
            & 0.600 & 0.300 & 0.275 & 0.682 & 0.000 & 0.000 \\
        \bottomrule
    \end{tabular}
    \caption{\textbf{Pre-SAPO group-wise attention--prediction accuracy}
    (mIoU). Queries are grouped by comparing the ground-truth IoU of TAD and
    PTD.}
    \label{tab:base_group_miou}
\end{table*}

\noindent\textbf{Post-SAPO TAD--PTD group analysis.}
We further partition evaluation queries according to whether the trained
policy's temporal attention decoding (TAD) or predicted timestamp
decoding (PTD) attains the higher IoU with the ground truth. Charades-STA uses
the full test set, while ActivityNet-Captions uses the same fixed 4,000-query
analysis subset as the preceding attention experiments. Table~\ref{tab:post_sapo_groups}
reports the number and proportion of examples in each group together with the
within-group mIoU of TAD and PTD.

\begin{table*}[t]
    \centering
    \footnotesize
    \setlength{\tabcolsep}{5pt}
    \begin{tabular}{llrrrr}
        \toprule
        \textbf{Dataset} & \textbf{Group} & \textbf{Samples} &
        \textbf{Share} & \textbf{TAD mIoU} & \textbf{PTD mIoU} \\
        \midrule
        \multirow{3}{*}{Charades-STA}
            & Attention better (TAD $>$ PTD) & 1,682 & 45.2\% & 0.565 & 0.352 \\
            & Prediction better (PTD $>$ TAD) & 1,745 & 46.9\% & 0.310 & 0.535 \\
            & Tie (TAD $=$ PTD) & 293 & 7.9\% & 0.000 & 0.000 \\
        \midrule
        \multirow{3}{*}{ActivityNet-Captions}
            & Attention better (TAD $>$ PTD) & 1,689 & 42.2\% & 0.454 & 0.250 \\
            & Prediction better (PTD $>$ TAD) & 1,957 & 48.9\% & 0.293 & 0.457 \\
            & Tie (TAD $=$ PTD) & 354 & 8.9\% & 0.000 & 0.000 \\
        \bottomrule
    \end{tabular}
    \caption{\textbf{Post-SAPO attention--prediction group statistics.}
    Queries are grouped by comparing the ground-truth IoU of TAD and PTD.
    TAD mIoU and PTD mIoU are computed within each group.}
    \label{tab:post_sapo_groups}
\end{table*}

Both attention-better and prediction-better examples remain substantial after
SAPO. The large within-group margins show that the near-closed aggregate
generation gap does not imply uniform agreement on every query; instead, the
two signals retain complementary strengths across different examples.

\subsection{SAPO Reward and Sample Selection}
\noindent\textbf{Distributions.}
A predicted interval $[s,e]$ is turned into a temporal distribution $g([s,e])$
by placing a Gaussian over the $T$ uniform frame bins, centered at $(s{+}e)/2$
with standard deviation $0.15\,\max(e{-}s,\Delta)$, where $\Delta$ is the bin
width, and then normalizing. The temporal distribution reward uses the
one-dimensional
Wasserstein-1 distance
$W_1(p,q)=\tfrac1T\sum_i|\mathrm{CDF}_p(i)-\mathrm{CDF}_q(i)|$, normalized to
$[0,1]$.

\noindent\textbf{Reward weights.}
We use $w_{\mathrm{dist}}=1.0$ for the temporal distribution reward,
$w_{\mathrm{IoU}}=0.5$ for the attention-interval IoU reward,
$\lambda_{\mathrm{fmt}}=0.1$ for the format-validity reward, and
$\lambda_{\mathrm{width}}=0.5$ for interval-width regularization
$p_{\mathrm{width}}=\max(0,(e{-}s)/\mathrm{dur}-0.5)$. Malformed or unparseable
outputs receive a fixed penalty of $-1$. Ground-truth timestamps are used only
for diagnostic analyses, the oracle upper bound, and final evaluation. They
never enter sample selection, reward construction, optimization, or model
selection in the standard SAPO setting.
\begin{algorithm}[h]
\caption{\method{} training}
\label{alg:sapo}
\begin{algorithmic}[1]
\STATE \textbf{Input:} training pool $\mathcal{D}$ without temporal boundaries, policy $\theta$, frozen reference $\theta_0$
\STATE run $\pi_{\theta_0}$ once; construct temporal attention $a$ and compute prediction--attention JS
\STATE $\mathcal{D}_{\mathrm{sel}}\leftarrow$ samples with the highest JS divergence
\FOR{each step}
    \STATE sample a batch from $\mathcal{D}_{\mathrm{sel}}$; draw a rollout group per query from $\pi_\theta$
    \STATE read reference attention distribution $a$ and interval $I_a$ from $\pi_{\theta_0}$
    \STATE score each rollout with the composite reward $r$
    \STATE update $\theta$ with group-relative advantage and KL to $\pi_{\theta_0}$
\ENDFOR
\end{algorithmic}
\end{algorithm}

\begin{table}[h]
    \centering
    \small
    \setlength{\tabcolsep}{2pt}
    \begin{tabular}{lcc}
        \toprule
        \textbf{Setting} & \textbf{Charades} & \textbf{ActivityNet} \\
        \midrule
        Qwen2.5-VL-3B + SAPO & 40.9 $\pm$ 0.3 & 32.8 $\pm$ 0.4 \\
        Qwen2.5-VL-7B + SAPO & 46.1 $\pm$ 0.2 & 33.3 $\pm$ 0.4 \\
        Time-R1-7B + SAPO    & 53.2 $\pm$ 0.2 & 46.7 $\pm$ 0.3 \\
        \midrule
        SAPO from test sets  & 42.0 $\pm$ 0.3 & 33.9 $\pm$ 0.4 \\
        \midrule
        Random selection     & 39.5 $\pm$ 0.3 & 31.1 $\pm$ 0.4 \\
        JS selection         & 40.9 $\pm$ 0.3 & 32.8 $\pm$ 0.4 \\
        Oracle selection     & 42.1 $\pm$ 0.2 & 34.1 $\pm$ 0.3 \\
        \bottomrule
    \end{tabular}
    \caption{\textbf{Run-to-run stability} (mIoU, mean $\pm$ std.).}
    \label{tab:run_stability}
\end{table}
\begin{table}[h]
    \centering
    \small
    \setlength{\tabcolsep}{5pt}
    \begin{tabular}{lc}
        \toprule
        \textbf{Model} & \textbf{Video-MME Acc.} \\
        \midrule
        Qwen2.5-VL-3B & 58.0 \\
        + \method{}   & 57.4 $\pm$ 0.3 \\
        \midrule
        Time-R1-7B    & 60.1 \\
        + \method{}   & 60.3 $\pm$ 0.2 \\
        \bottomrule
    \end{tabular}
    \caption{\textbf{General-ability stability across SAPO runs}
    (accuracy, mean $\pm$ std.).}
    \label{tab:videomme_stability}
\end{table}

\begin{table}[h]
    \centering
    \small
    \setlength{\tabcolsep}{1.2pt}

    \begin{tabular}{lcccc}
        \toprule
        \textbf{Model} & \textbf{Overall} & \textbf{Short} & \textbf{Medium} & \textbf{Long} \\
        \midrule
        Qwen2.5-VL-3B-Instruct & 58.0 & 65.7 & 58.6 & 49.9 \\
        + \method{} on Charades & 57.0 & 65.2 & 57.2 & 48.7 \\
        + \method{} on ActivityNet & 57.4 & 65.7 & 57.8 & 48.7 \\
        + \method{} on Charades+ActivityNet & 57.4 & 66.2 & 57.2 & 48.7 \\
        \midrule
        Time-R1-7B & 60.1 & 69.7 & 60.3 & 50.2 \\
        + \method{} on Charades & 60.4 & 69.9 & 60.4 & 51.0 \\
        + \method{} on ActivityNet & 60.3 & 70.0 & 60.4 & 50.6 \\
        + \method{} on Charades+ActivityNet & 60.3 & 70.0 & 60.4 & 50.3 \\
        \bottomrule
    \end{tabular}
    \caption{\textbf{Full Video-MME results} (accuracy, \%).}
    \label{tab:videomme_full}
\end{table}

\noindent\textbf{Selection.}
We rank the training pool without temporal boundaries by prediction--attention
disagreement, measured as
the Jensen--Shannon divergence between $g(I_p)$ and the corrected attention
distribution $a$; larger values indicate higher training value. The main pool
is constructed according to the data source. For single-source adaptation, we
select 2,000 examples from Charades-STA or 2,000 examples from
ActivityNet-Captions. For joint adaptation, we select 1,000 examples from each
dataset, resulting in the same total budget of 2,000 examples. We use these
budgets consistently whether candidates are drawn from the training split or,
in the transductive setting, from the test split.

\subsection{Training Configuration}
We train all base models with GRPO under the verl framework on five GPUs in a
colocated actor/reference/rollout configuration, using sglang for rollout
generation. The grounding prompt asks the model to output the start and end
times of the queried event in seconds. Unless noted below, the three base
models share the same optimizer settings: a learning rate of
$1\!\times\!10^{-6}$, KL coefficient $\beta=0.03$ to the frozen reference model,
entropy coefficient $0.01$, a maximum prompt length of 4,096, and two epochs.
We monitor the SAPO training reward and select a checkpoint after
the reward has converged. No ground-truth or labeled-validation metric is used
for checkpoint selection.

For \textbf{Qwen2.5-VL-3B-Instruct} and \textbf{Qwen2.5-VL-7B-Instruct}, each
step draws a training batch of 60 queries with a PPO mini-batch of 30 and
samples eight rollouts per query at temperature $1.1$, with a maximum response
length of 48 tokens (the model emits a short \texttt{[s,\,e]} timestamp). The
two scales share the same configuration and differ only in the frozen base
model and its representative attention layer (Table~\ref{tab:attn_layers}).

For \textbf{Time-R1-7B}, which produces chain-of-thought before its answer, we
use a training batch of 30 queries with a PPO mini-batch of 15, eight rollouts
per query at temperature $1.0$, and a maximum response length of 256 tokens to
accommodate the longer reasoning trace. All other hyperparameters are identical
to the shared settings above.

\subsection{Run-to-Run Stability}
We repeat the principal SAPO settings with three random seeds.
Table~\ref{tab:run_stability} reports mIoU as mean and standard deviation. The
main-model rows use joint training-set adaptation, the transductive row uses the
joint test-set pool, and the selection rows use the matched mixed-data budget
from Table~\ref{tab:ablation_select}.

\subsection{General Video Understanding}
The following tables report run-to-run stability and results by video duration
for the Video-MME evaluation in the main paper.

\end{document}